\documentclass[lettersize,journal]{IEEEtran}
\usepackage{amsmath,amsfonts}
\usepackage{algorithmic}
\usepackage{algorithm}
\usepackage{array}

\usepackage[caption=false,font=normalsize,labelfont=sf,textfont=sf]{subfig}
\usepackage{amsmath}
\usepackage{textcomp}
\usepackage{stfloats}
\usepackage{url}
\usepackage{verbatim}
\usepackage{graphicx}
\usepackage{multirow}
\usepackage{booktabs}
\usepackage{enumerate}

\usepackage{cite}
\begin{document}

\title{Dual-Branch Residual Network for Cross-Domain Few-Shot Hyperspectral Image Classification with Refined Prototype}

\author{Anyong Qin, Chaoqi Yuan, Qiang Li, Feng Yang, Tiecheng Song and Chenqiang Gao
\thanks{Manuscript received April 19, 2021; revised August 16, 2021. 
This work was supported by the National Natural Science Foundation of China under Grant 62201111; in part by the China Postdoctoral Science Foundation under Grant 2024T171109 and Grant 2022MD713684; in part by the Special Funding for Chongqing Postdoctoral Research Project under Grant 2022CQBSHTB3086. (Corresponding author: Tiecheng Song, songtc@cqupt.edu.cn)}

\IEEEcompsocitemizethanks{
\IEEEcompsocthanksitem Anyong Qin, Chaoqi Yuan, Qiang Li, Feng Yang and Tiecheng Song are with the School of Communications and Information Engineering, Chongqing University of Posts and Telecommunications, Chongqing 400065, China.
\IEEEcompsocthanksitem Chenqiang Gao is with the School of Intelligent Systems Engineering, Sun Yat-sen University, Shenzhen, Guangdong 518107, China.\protect\\
}
}

\markboth{IEEE Geoscience and Remote Sensing Letters~2025}%
{Shell \MakeLowercase{\textit{et al.}}: A Sample Article Using IEEEtran.cls for IEEE Journals}
\maketitle

\begin{abstract}
Convolutional neural networks (CNNs) are effective for hyperspectral image (HSI) classification, but their 3D convolutional structures introduce high computational costs and limited generalization in few-shot scenarios. Domain shifts caused by sensor differences and environmental variations further hinder cross-dataset adaptability. Metric-based few-shot learning (FSL) prototype networks mitigate this problem, yet their performance is sensitive to prototype quality, especially with limited samples. To overcome these challenges, a dual-branch residual network that integrates spatial and spectral features via parallel branches is proposed in this letter. Additionally, more robust refined prototypes are obtained through a regulation term. Furthermore, a kernel probability matching strategy aligns source and target domain features, alleviating domain shift. Experiments on four publicly available HSI datasets illustrate that the proposal achieves superior performance compared to other methods.
\end{abstract}


\section{Introduction}
\IEEEPARstart{H}{yperspectral} image (HSI) classification aims to accurately identify land cover types by analyzing spectral features and plays a crucial role in agricultural and environmental monitoring. 

Approaches utilizing hand-crafted features and shallow classifiers have demonstrated a limited capacity to extract complex spectral-spatial information effectively. Convolutional neural networks (CNNs), a type of deep learning technique, have greatly improved classification accuracy by automatically learning effective features \cite{3dcnn-li2017spectral,ssrn-zhong2017spectral}. However, most CNN-based models assume that the training and test data share similar distributions, which is rarely true in real-world scenarios due to sensor discrepancies and environmental variations. This domain shift reduces the generalization performance of supervised models. To address this problem, domain adaptation (DA) approaches have been proposed, which can be generally classified into those based on adversarial strategies \cite{dann} and those based on discrepancy minimization \cite{long2013mmd}. 
Discrepancy-based approaches, such as those using maximum mean discrepancy (MMD) \cite{long2013mmd}, offer a simpler, more stable alternative. Despite their effectiveness, misalignment and performance degradation due to negative transfer, especially when the target domain includes unseen categories.

Metric-based FSL offers a promising solution using knowledge from base classes to classify novel categories with limited samples \cite{DFSL,CMFSL,luo2024sdst,luo2023multiscale,qin2024few,qin2024deep}. DFSL \cite{DFSL} introduced a prototype network that classifies samples based on Euclidean distances to class prototypes. 
CMFSL \cite{CMFSL} incorporated class covariance metrics and spectral-prior-based refinement to reduce overfitting and domain shift. However, class prototypes are easily influenced by outliers when the support set is limited (e.g., shot=1), causing deviations from the true class centers.
To address this challenge, researchers have proposed variants of prototype networks to enhance prototype representativeness. SDST \cite{luo2024sdst} leverages a shared autoformer for global feature learning and a Siamese dual-former graph module for fine-grained local detail extraction, while Qin et al. \cite{qin2024few} leveraged query set information to refine unstable prototypes. RPCL-FSL \cite{rpcl} adopted triplet loss to improve prototypes. The complex spectral-spatial characteristics in HSI data are often difficult to fully capture with existing feature extraction networks.
Moreover, adversarial-based approaches such as CA-CFSL \cite{ca-cfsl} suffer from slow convergence, training instability, and complex loss functions.

To address these challenges, this letter presents a cross-domain few-shot HSI classification framework based on a dual-branch residual network. The network integrates spatial and spectral features through parallel branches and fuses them within residual blocks, improving class separability and feature robustness, while reducing computational complexity. To address prototype instability, we introduce contrastive constraints inspired by contrastive learning, which improve class discrimination by pulling intra-class features closer and increasing the separation between inter-class distributions. For domain shift, we use MMD to align source and target features, ensuring stable and efficient adaptation without the convergence issues of adversarial methods.

\section{Proposed Method}
\label{Proposed Method}
\begin{figure}[!t]
\centering
\resizebox{\linewidth}{!}
{
\includegraphics{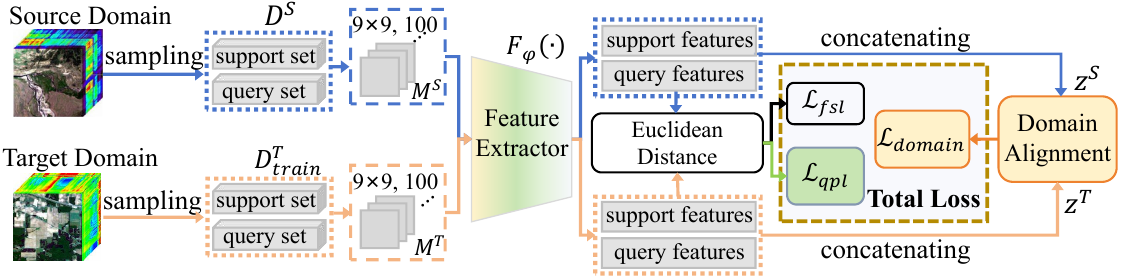}
}
\caption{Architecture of the proposed method, which includes four main modules. First, the mapping layers are employed to unify the dimensionality of $D^S$ and $D^T$ after the data pre-processing. Second, a dual-branch feature extractor captures spatial and spectral features independently, followed by feature fusion for enhanced representation. $F_\psi(\cdot)$ represents the feature extractor with parameters $\psi$. Third, to obtain more robust refined prototypes, we use the Query-Prototype contrastive refinement Loss (QPL) to improve inter-class separability and minimize intra-class variation. Finally, the domain alignment module leverages maximum mean discrepancy (MMD) to align feature distributions, reducing domain shifts in HSI. Training alternates between source and target domains. When a training episode is completed on either source domain or target domain. The source domain total loss $L^s_{total}$(the target domain total loss $L^t_{total}$) will be back propagated to update the feature extractor parameters.}
\label{fig:framework}
\end{figure}

In this section, we give a detailed description of each module. The overall framework of training phase is shown in Figure \ref{fig:framework}.

\begin{figure}[!t]
\centering
\resizebox{\linewidth}{!}
{
\includegraphics{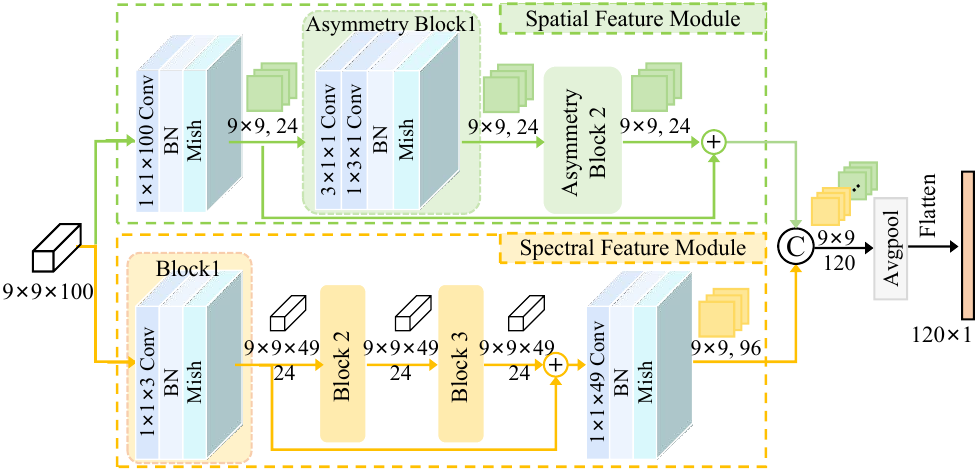}
}
\caption{Overview of the feature extraction module: A dual-branch network that extracts spatial and spectral features separately. The spatial branch uses asymmetric convolutions to capture spatial correlations, while the spectral branch employs layered convolutions to learn spectral patterns. Finally, the spatial and spectral features are fused, producing a 120-dimensional vector for the FSL stage.}
\label{fig:fe}
\end{figure}

\subsection{Dual-Branch Feature Extractor}
The role of the dual-branch feature extractor is to extract embedded features from the input data, which is structured with a mapping layer and a feature extractor.

Dimensional alignment is achieved by a mapping layer, which is architecturally a 2D convolutional layer, with \(d_{out}\) kernels of size \(1 \times 1 \times d_{in}\), where \(d_{in}\) and \(d_{out}\) denote the input and output channels, respectively. This layer unifies high-dimensional HSI cubes from the source domain ($SD$) and the target domain ($TD$) into representations of \(9 \times 9 \times 100\).

HSI data contain rich spectral and spatial information, posing challenges for traditional 3D CNN-based feature extraction methods. Processing spatial features $\mathbf{F}_{sp}$ and spectral features $\mathbf{F}_{spc}$ simultaneously can blur domain-specific patterns and increase computational complexity. Separating them enables more focused feature learning and mitigates feature interference. Therefore, to effectively exploit the spatial-spectral information in HSI, we designed a dual-branch feature extractor that extracts $\mathbf{F}_{sp}$ and $\mathbf{F}_{spc}$ separately and subsequently fuses them. The proposed network structure is illustrated in Figure \ref{fig:fe}.

\subsubsection{Spatial Feature Extraction}
As illustrated in the upper branch of Figure \ref{fig:fe}, this module integrates a convolutional layer and a residual block. The first convolutional layer is designed to learn spatial correlations. 
The residual block comprises two asymmetric convolutional layers with kernel sizes of \((3,1,1)\) and \((1,3,1)\), respectively. Subsequent to each convolution step, batch normalization (BN) and the Mish activation \cite{mish} are implemented. This approach to asymmetric convolution is intended to minimize the quantity of network parameters, while preserving and emphasizing critical spatial information.

\subsubsection{Spectral Feature Extraction}
This module is built with two convolutional layers and a residual block, which is designed to preserve spatial correlations while effectively representing the spectral information. As shown in the lower branch of Figure \ref{fig:fe}, the residual block includes two convolutional layers. Similar to the spatial module, each convolution is followed by a BN layer and the Mish activation function. The residual block further enhances the robustness of the $\mathbf{F}_{spc}$ while maintaining the same feature map size. 

\subsubsection{Spatial-Spectral Feature Fusion}
The concatenated features form a 120-channel representation with spatial dimensions of \((9,9)\). The concatenation operation retains complementary information from both branches, ensuring that spatial patterns and spectral correlations are jointly represented. An average pooling operation with a kernel size of \((9,9)\) is applied to produce a 120-dimensional vector. The resulting vector is then passed to the FSL stage.

\subsection{Data Pre-processing and Construction of FSL task}
Both the source domain and target domain data were pre-processed before training. For the source domain (Chikusei), classes with less than 200 samples were removed, leaving 18 classes. From each class, 200 samples were randomly selected. For the target domain, 5 samples were randomly selected from each class. These were then augmented with random Gaussian noise to expand each class to 200 samples. The remaining samples  from the target domain were used for testing. Details of the pre-processing and data augmentation can be found in related work \cite{dcfsl,rpcl,CMFSL}.

For episodic FSL training, tasks are formed by independently sampling samples from both $SD$ and $TD$ datasets. To illustrate the episodic training process, we describe the task construction in $SD$ as a demonstrative case. The first step involves randomly picking \(C\) categories from the $SD$ classes. Next, from every chosen category, \(N_s\) samples are randomly selected from $D^S$ for the support set, and \(N_q\) samples for the query set. It's important to point out that the samples within the support set are different from those in the query set. Each episodic task contains \(C\times(N_s + N_q)\) samples.

The calculation formula for the $c$-th prototype $\mu_c$ is as follows:
\begin{equation}
\label{eq:compute_prototype}
\mu_c = \frac{1}{N_s} \sum_{z_i \in S_c} F_\psi(z_i)
\end{equation}
where $z_i$ denotes the sample of the \(c\)-th class in the support set $S_c$. $F_\psi(\cdot)$ represents the feature extractor with parameters $\psi$.

The probability prediction formula for a sample $z_q$ in the query set belonging to class \(c\) is:
\begin{equation}
\label{eq:compute_p}
\Gamma_{\psi}(y_q = c | z_q \in Q) = \frac{\exp(-ED(F_\psi(z_q), \mu_c))}{\sum_{c=1}^{C}\exp(-ED(F_\psi(z_q), \mu_{c}))}
\end{equation}
where $Q$ represents the query set, $z_q$ is the query sample, and $ED(\cdot, \cdot)$ denotes the Euclidean distance.

The FSL loss on $SD$ can be computed using the true label and the negative log probability of the query samples as:
\begin{equation}
\label{eq:compute_source_fsl}
L_{fsl}^s = \mathbb{E}_{(S_s,Q_s)} \left[ -\sum_{(z,y)\in Q_s}\log {\Gamma}_{\psi}(y=c|z)\right] 
\end{equation}

Similarly, the FSL loss on the $TD$ dataset can be calculated using the subsequent formula:
\begin{equation}
\label{eq:compute_target_fsl}
L_{fsl}^t = \mathbb{E}_{(S_t,Q_t)} \left[ -\sum_{(z,y)\in Q_t}\log {\Gamma}_{\psi}(y=c|z)\right] 
\end{equation}

\subsection{Learning Refined Prototypes with QPL}
Prototype networks classify query samples using class prototypes but face two main limitations: prototypes obtained by direct averaging of support set features are prone to noise in the few-shot scenarios. The classification process relies solely on support set prototypes without effectively leveraging query set label information. 

To effectively address the problems discussed above, a refinement strategy Query-Prototype contrastive refinement Loss (QPL) based on comparative learning of query samples and prototypes is proposed. By limiting the distance between query samples and class prototypes, the method simultaneously improves class separation and intra-class compactness in the feature space. We calculate the contrast loss for each query sample against all prototypes of different classes (negative prototypes) as follows:
\begin{equation}
\label{eq:inter_class_loss}
L_{inter} = \frac{1}{(C-1)CN_{q}}\sum_{\substack{i,j=1 \\ j\neq i}}^{C}\sum_{k=1}^{N_q}log(1+e^{-ED(f_{i,k}^{q},\mu_{j}^{s})})
\end{equation}

Similarly, we will compute the loss of contrast between each query sample and the positive prototype to enhance the intra-class compactness as follows:
\begin{equation}
\label{eq:intra_class_loss}
L_{intra} = \frac{1}{CN_q}\sum_{i=1}^{C}\sum_{k=1}^{N_q}log(1+e^{ED(f_{i,k}^{q},\mu_{i}^{s})})
\end{equation}
where $\mu_i^s$ is defined as the prototype of the $i$-th class, while $f_{i,k}^{q}$ corresponds to the feature vector of the $k$-th query sample from the $i$-th class in the query set. 

Thus, the final loss function \(L_{qpl}\) is computed as: 
\begin{equation}
\label{eq:qpl_loss}
L_{qpl} = L_{inter} + L_{intra}
\end{equation}

Unlike traditional contrastive losses such as Triplet Loss, QPL does not require extensive experiments to fine-tune the margin parameter \(\delta\). Additionally, the logarithmic form of QPL enhances gradient stability and avoids excessive penalties for misclassified samples. By optimizing QPL, the intra-class clustering is made more compact while reducing the inter-class similarity, resulting in a more robust refined prototype.

\subsection{Domain Alignment}
Cross-domain generalization faces the challenge of distribution shifts caused by differences in geographic environments and sensor characteristics. MMD is recognized as a valuable non-parametric metric for quantifying these distributional differences, and it is frequently employed to bridge the gap between $D^S$ and $D^T$, and achieve domain alignment. The detailed calculation process is outlined below:
\begin{equation}
\label{eq:domain_loss}
L_{domain} = \left\Vert \frac{1}{N_{S}}\sum_{i=1}^{N_S}\phi(z_i^S) - \frac{1}{N_{T}}\sum_{j=1}^{N_T}\phi(z_j^T) \right\Vert_{H}^{2}
\end{equation}
where \(\phi\) represents the transformation that maps samples into a reproducing kernel Hilbert space. During training on $SD$, \(z^{S}\) is formed by concatenating query and support features from $SD$. \(z^{T}\) is randomly sampled from \(D^T\).  
Similarly, during training on $TD$, \(z^{T}\) is formed by concatenating query and support features from $TD$. \(z^{S}\) is randomly sampled from \(D^S\). \(N_S\) and \(N_T\) as the sample size of \(z^{S}\) and \(z^{T}\). 
This metric measures the average discrepancy between $D^S$ and $D^T$, with a smaller $L_{domain}$ value indicating greater similarity between the two domains.

The total loss $L_{total}^s$ on $SD$ is expressed as:
\begin{equation}
\label{eq:source_total_loss}
L_{total}^s = L_{fsl}^s+L_{qpl}+L_{domain}
\end{equation}

Similarly, the final loss function $L_{total}^t$ on $TD$ is computed as:
\begin{equation}
\label{eq:target_total_loss}
L_{total}^t = L_{fsl}^t+ L_{qpl}+L_{domain}
\end{equation}

In conclusion, the complete loss $L_{total}$ for the proposed method is denoted as:
\begin{equation}
\label{eq:total_loss}
L_{total} = L_{total}^s+L_{total}^t
\end{equation}

In (\ref{eq:total_loss}), the losses in the source and target domains are weighted equally. This helps the model to learn from the source domain while adapting to the target domain.
\section{EXPERIMENTS}
\label{EXPERIMENRS}
\subsection{Datasets and Experimental setting}
\subsubsection{HSI Datasets}
In the experiments, we select Chikusei dataset as the source domain dataset. Indian Pines (IP), Salinas (SA), and Pavia University (UP), and Houston 2013 (HS) as the target domain datasets. 
\textbf{Chikusei}: The Chikusei dataset, acquired in Chikusei using hyperspectral visible/near-infrared cameras, includes 19 different categories. It has a spatial resolution of 2517 × 2335 pixels and contains 128 spectral bands.
\textbf{IP}: The IP dataset was acquired by AVIRIS sensors, contains $145 \times 145$ pixels, with 10,249 labeled pixels divided into 16 classes and 200 spectral bands. 
\textbf{SA}: It was captured by AVIRIS, includes $512 \times 217$ pixels. It contains 54,129 labeled pixels distributed over 16 land cover classes and 204 spectral bands.
\textbf{UP}: The UP dataset was collected by ROSIS, contains $610 \times 340$ pixels, with 42,776 labeled pixels in 9 classes and 103 spectral bands.
\textbf{HS}: The Houston 2013 dataset includes 15,029 pixels, which had fixed training and test sets corresponding to sample sizes of 2,832 and 12,197 respectively. It contains 15 categories and 145 spectral bands.
\subsubsection{Experimental Setting}
The input cube has a spatial size of 9 × 9. The training strategy is based on episodic training, lasting 10,000 episodes and uses the Adam optimizer with a learning rate of $0.001$. 
In our method, $N_{s}$, \(N_q\) and \(L\) was set to $1$, $19$ and $5$ respectively. For the IP, SA, UP and HS datasets, $C$ was set to 16, 16, 9 and 15, respectively.
The Overall Accuracy (OA), Average Accuracy (AA), and Kappa coefficient served as the evaluation metrics for all methods' performance. To eliminate the effect of random sampling, experiments were run $10$ times on the four target datasets. The highest and second accuracies in each row are bolded and underlined respectively.

\subsection{Methods Comparison and Analysis}
The comprehensive comparisons were conducted to validate the effectiveness of the proposed method, including 3D-CNN\cite{3dcnn-li2017spectral}, SSRN\cite{ssrn-zhong2017spectral}, DFSL+NN\cite{DFSL}, DCFSL\cite{dcfsl}, Gia-CFSL\cite{gia-cfsl}, CMFSL\cite{CMFSL}, RPCL-FSL\cite{rpcl}, DACAA\cite{DACAA} and FDFSL\cite{FDFSL}. 
Classification results for all methods on the IP, SA, UP and HS datasets are shown in Table \ref{tab:classification_res_on_four_datasets}.
\begin{table*}[!ht]
\caption{Classification Performance (\%) of Various Methods on Four Target Domain Datasets 
\label{tab:classification_res_on_four_datasets}}
\centering
\resizebox{16.0cm}{!}
{
\begin{tabular}{cccccccccccc}
\hline \toprule[0.8pt]
Dataset             & Metric &3D-CNN\cite{3dcnn-li2017spectral} & SSRN\cite{ssrn-zhong2017spectral}     &  DFSL+NN\cite{DFSL}&DCFSL\cite{dcfsl}       &Gia-CFSL\cite{gia-cfsl}& CMFSL\cite{CMFSL}     & RPCL-FSL\cite{rpcl}    &  DACAA\cite{DACAA}&FDFSL\cite{FDFSL}& Proposed            \\ \hline
\multirow{3}{*}{IP} & OA     &56.04±5.00& 61.17±1.41  &  59.47±1.86&65.03±2.21  &64.26±2.60& 66.47±3.10 & \underline{73.85±2.74}  &  68.08±2.59&69.64±2.19& \textbf{74.83±2.33} \\
                    & AA     &70.76±3.89& 76.86±2.51 &  71.41±1.23&76.53±1.45  &75.68±1.50& 79.10±1.56 & \underline{83.11±1.99}  &  79.49±1.43&80.66±1.92& \textbf{84.05±2.09} \\
                    & Kappa  &50.77±5.34& 56.84±1.68  &  54.60±1.92&60.60±2.26  &59.68±2.75& 62.32±3.21 & \underline{70.62±2.19}  &  64.03±2.75&65.90±2.40& \textbf{71.62±2.34} \\ \hline
\multirow{3}{*}{SA} & OA     &82.18±4.82& 86.87±2.58 &  85.55±1.19&89.62±1.43  &88.90±1.62& 89.13±1.84 & \underline{90.89±1.00}  &  89.04±1.85&90.92±0.68& \textbf{91.68±1.18} \\
                    & AA     &87.56±3.07& 91.25±1.64  &  90.62±0.96&94.07±1.09  &92.81±1.48& 94.36±0.82 & 94.38±1.11  &  93.12±1.09&\underline{94.61±0.91}& \textbf{95.23±1.08} \\
                    & Kappa  &80.20±5.31& 85.40±2.86  &  83.98±1.29&88.47±1.55  &87.67±1.77& 87.94±2.02 & \underline{89.94±1.10}  &  87.73±2.02&89.91±0.75& \textbf{90.74±1.30} \\ \hline
\multirow{3}{*}{UP} & OA     &66.29±5.18& 77.24±3.19  &  76.44±2.57&81.30±3.09  &81.97±2.56& 80.56±4.65 & 81.51±3.37  &\underline{83.07±2.76}  &80.89±2.78& \textbf{84.72±2.27} \\
                    & AA     &75.58±3.26& 80.41±3.06 &  77.22±2.39&81.31±1.20  &81.81±1.83& 83.10±2.24 & 84.79±2.33  &\underline{85.40±2.28}  &82.64±2.02& \textbf{86.33±2.92} \\
                    & Kappa  &58.35±5.69& 70.64±3.94 &  69.67±2.87&75.75±3.57  &76.58±3.09& 75.24±5.37 & 76.46±3.86  &\underline{78.07±3.38}  &75.21±3.34& \textbf{80.22±2.79} \\ \hline
\multirow{3}{*}{HS} & OA     &66.70±2.31& 68.55±4.47  &  -         &74.37±2.89  &73.15±2.47&  -         & 74.61±2.29  &\underline{77.05±1.64}  &76.89±2.45&  \textbf{78.49±2.56}\\
                    & AA     &67.51±2.17&73.02±3.58  &  -         &76.29±2.24  &74.81±2.15&  -         & 77.65±1.91  &79.41±1.51  &\underline{79.43±1.76}&  \textbf{80.46±1.92}\\
                    & Kappa  &63.51±2.52&65.93±4.79  &  -         &72.22±3.12  &70.90±2.65&  -         & 72.50±2.48  &\underline{75.28±1.77}  &74.96±2.62&  \textbf{76.67±2.77}\\ \bottomrule[0.8pt]\hline

\end{tabular}
}
\end{table*}
FSL methods outperform traditional deep learning approaches such as 3D-CNN and SSRN in adaptability and generalization. Among FSL methods, DFSL performs poorly due to its inability to handle domain shifts in datasets. DCFSL improves performance by integrating cross-domain training, while Gia-CFSL uses graph structures to enhance feature extraction and generalization. CMFSL optimizes metric learning for decision boundaries and reduces misclassification, while FDFSL employs an orthogonal low-rank disentangling method to separate features and encourage domain-specific information learning. However, none of these methods consider both feature extraction and prototype quality simultaneously. In contrast, our proposed method enhances prototype robustness and reduces classification confusion, consistently outperforming other methods across all datasets.
\begin{figure}[!htbp]
\centering
\resizebox{\linewidth}{!}
{
\includegraphics{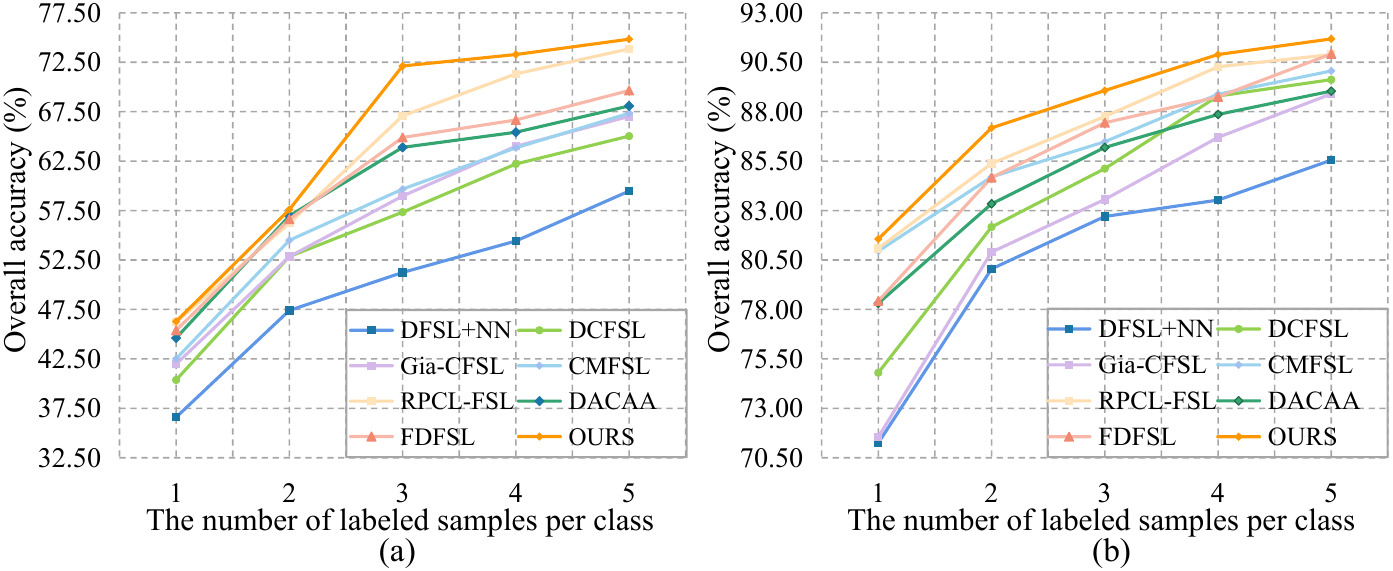}
}
\caption{Classification result of different methods on the four data sets with different number of labeled samples. (a) IP. (b) SA.}
\label{fig:different_numbers_of_labeled_samples}
\end{figure}

To verify the effect of the number of labeled samples on the performance of the model, we randomly selected \(L\) labeled samples from each class of each target domain dataset to observe the OA of the different methods. Figure \ref{fig:different_numbers_of_labeled_samples} illustrates the OA variations of IP and SA as the number of labeled samples increases. As the number of available labeled samples increases, the OA of all models is improved. 
Notably, our proposal consistently demonstrates superior classification performance compared to other methods.

\subsection{Ablation Study}
To further analyze the contribution of each component, ablation studies were conducted to isolate the effects of individual modules and network configurations.
\begin{table}[!ht]
\caption{Classification Results(OA\%) with Different Modules
\label{tab:classification_res_ablation}}
\centering
\resizebox{\linewidth}{!}
{
\begin{tabular}{ccccccc}
\hline \toprule[0.8pt]
\multicolumn{3}{c}{Method}                    & \multicolumn{4}{c}{Dataset}\\ \cline{4-7} 
Feature Extractor                                & QPL         & MMD          & IP           & SA           & UP             &HS\\ \hline
\multirow{4}{*}{FE1} & ×             & ×           & 65.03±2.21   & 89.62±1.43   & 81.30±3.09     &74.37±2.89\\
                     & \checkmark    & ×           & 69.46±2.43   & 90.51±1.56   & 82.22±2.83     & 75.01±2.41\\
                     & ×             & \checkmark  & 66.11±1.89   & 90.03±1.34   & 81.99±2.76     &74.72±2.20\\
                     & \checkmark    & \checkmark  & 71.60±2.77   & 90.99±1.21   & 82.79±2.02     &75.48±2.53\\ \hline
\multirow{4}{*}{FE2} & ×             & ×           & 73.66±2.37   & 90.59±2.01   & 83.45±2.78     &77.13±2.39\\
                     & \checkmark    & ×           & 74.17±2.48   & 91.04±1.78   & 84.03±2.91     &77.87±3.61\\
                     & ×             & \checkmark  &  74.03±2.12  & 91.12±1.45   & 84.12±2.40     &77.81±2.82\\
                     & \checkmark    & \checkmark  & 74.83±2.33   & 91.68±1.18   & 84.72±2.27     &78.49±2.56\\ \bottomrule[0.8pt] \hline
\end{tabular}
}
\end{table}
\subsubsection{The effectiveness of the different modules}
DCFSL \cite{dcfsl} is used as the baseline, and the feature extractor used in DCFSL is denoted as FE1. The proposed network is denoted as FE2. Specifically, each module is added individually to both FE1 and FE2, and the classification results before and after adding each module are compared to evaluate its performance. 
Table \ref{tab:classification_res_ablation} presents the experimental results of applying the QPL and MMD modules individually or jointly to FE1 (Rows 1–4) and FE2 (Rows 5–8). Comparing Row 1 and Row 5, the proposed feature extractor improves the OA by 8.63\%, 0.97\%, 2.15\% and 2.76\% on IP, SA, UP and HS, respectively, demonstrating stronger feature representation and robustness under few-shot conditions. Furthermore, comparing Rows 2-4 with Rows 5-8 shows that adding either the loss module or MMD alone or combining both leads to different degrees of improvement in OA. These results confirm the effect of these modules.
\begin{table}[!ht]
\caption{Classification Performance (\%) by Activation Function and Convolution Block 
\label{tab:classification_res_on_conv_block}}
\centering
\resizebox{7.7cm}{!}
{
\begin{tabular}{cccccc}
\hline \toprule[0.8pt]
\multirow{2}{*}{Datasets} & \multirow{2}{*}{Metrics} & \multicolumn{2}{c}{Activation Function} & \multicolumn{2}{c}{Convolution block} \\ \cline{3-6} 
                          &             & ReLU            & Mish             & Asym            & Sym     \\ \hline
\multirow{3}{*}{IP}       & OA          & 74.22±3.36      & 74.83±2.33       & 74.83±2.33      & 74.24±2.87       \\
                          & AA          & 83.55±3.30      & 84.05±2.09       & 84.05±2.09      & 83.11±1.78       \\
                          & Kappa       & 70.99±3.65      & 71.62±2.34       & 71.62±2.34      & 71.05±3.12       \\ \hline
\multirow{3}{*}{SA}       & OA          & 91.16±1.57      & 91.68±1.18       & 91.68±1.18      & 91.43±1.29 \\
                          & AA          & 94.76±1.24      & 95.23±1.08       & 95.23±1.08      & 94.17±0.88 \\
                          & Kappa       & 90.15±1.49      & 90.74±1.30       & 90.74±1.30      & 90.21±1.42 \\ \hline
\multirow{3}{*}{UP}       & OA          & 83.61±3.77      & 84.72±2.27       & 84.72±2.27      & 84.59±3.48       \\
                          & AA          & 84.84±2.46      & 86.33±2.92       & 86.33±2.92      & 85.60±1.91       \\
                          & Kappa       & 78.85±4.49      & 80.22±2.79       & 80.22±2.79      & 80.09±4.08       \\ \hline
 \multirow{3}{*}{HS}      & OA          & 78.33±3.26      & 78.49±2.56       & 78.49±2.56      & 78.31±2.41 \\
                          & AA          &80.21±1.91       & 80.46±1.92       & 80.46±1.92      & 80.36±1.77       \\
                          & Kappa       &76.54±2.56       & 76.67±2.77       & 76.67±2.77      & 76.56±2.61  \\ \bottomrule[0.8pt] \hline

\end{tabular}
}
\end{table}
\subsubsection{The impact of the feature extractor modules}
To further analyze the contribution of each module within the feature extraction network to model performance, systematic ablation experiments were conducted to evaluate module influence on four datasets. We tested different activation functions (ReLU and Mish) and different convolution block types (symmetric and asymmetric). The results are shown in Table \ref{tab:classification_res_on_conv_block}.

a) Activation Function: The experimental results are shown in columns 1 and 2 of Table \ref{tab:classification_res_on_conv_block}. Compared with ReLU, the Mish activation function achieves better performance on all four datasets, improving OA by 0.61\%, 0.52\%, 1.11\% and 0.16\%, respectively. This is due to the Mish being better at capturing the nonlinear relationships of the features when dealing with limited samples, thus improving the model's generalization and classification performance.

b) Asymmetric Convolution Block: The experimental results are presented in columns 3-4 of Table \ref{tab:classification_res_on_conv_block}. It is particularly noticeable in the IP dataset, with a 0.94\% improvement in AA. 
Classification accuracy results demonstrate the superiority of the asymmetric convolution block over the symmetric convolution block. By more effectively capturing the spatial and spectral features of the HSI, asymmetric convolution blocks enable models to achieve better generalization and classification performance with fewer parameters.

\subsection{Computational Complexity}
We compared the running time, FLOPs, and trainable parameters of different FSL algorithms, as shown in Table \ref{tab:table_complexity}. Despite a longer training phase, our method achieves superior classification results with a similar testing time compared to others.
\begin{table}[!ht]
\caption{Analysis of Parameters, Computational Time, and FLOPs for Various Methods Evaluated on Four Target Domain Datasets
\label{tab:table_complexity}}
\centering
\resizebox{\linewidth}{!}
{
\begin{tabular}{cccccccccc}
\hline \toprule[0.8pt]
\multicolumn{2}{c}{Method}         & DFSL+NN & DCFSL   & Gia-CFSL & CMFSL  & RPCL-FSL & DACAA  & FDFSL  & Proposed    \\ \hline
\multirow{4}{*}{IP} & Training (s) & 350.03  & 1351.40 & 3582.20  & 964.97 & 443.16   &3490.26 &1103.08 & 2483.41 \\
                    & Testing (s)  & 1.48    & 1.65    & 1.46     & 0.62   & 1.49     &6.18    &3.38    & 1.57    \\
                    & Params (M)   & 0.03    & 4.26    & 0.31     & 0.77   & 0.07     &1.12    &0.19    & 0.13    \\
                    & FLOPs (M)    & 41.51   & 47.35   & 41.78    & 29.57  & 43.18    &60.70   &52.72   & 25.76   \\ \hline
\multirow{4}{*}{SA} & Training (s) & 1066.40 & 1427.94 & 3731.00  & 994.19 & 543.96   &3642.39 &981.80  & 2559.73 \\
                    & Testing (s)  & 7.37    & 7.90    & 7.99     & 3.18   & 7.87     &24.89   &15.66   & 8.36    \\
                    & Params (M)   & 0.03    & 4.26    & 0.31     & 0.77   & 0.07     &1.12    &0.19    & 0.13    \\
                    & FLOPs (M)    & 41.51   & 47.38   & 41.78    & 29.57  & 43.21    &60.70   &52.76   & 25.76  \\ \hline
\multirow{4}{*}{UP} & Training (s) & 481.66  & 845.46  & 2468.35  & 654.87 & 310.90   &2815.50 &630.59  & 1441.25  \\
                    & Testing (s)  & 11.43   & 5.45    & 5.39     & 2.55   & 6.30     &22.11   &10.51   & 5.95    \\
                    & Params (M)   & 0.03    & 4.25    & 0.31     & 0.77   & 0.06     &1.12    &0.18    & 0.13    \\
                    & FLOPs (M)    & 41.51   & 46.57   & 41.78    & 29.57  & 42.39    &60.70   &51.94   & 25.76   \\  \hline
\multirow{4}{*}{HS} & Training (s) &-        & 1285.67 & 3274.82  &-       & 399.37   &3457.69 &777.89  & 2260.90      \\
                    & Testing (s)  &-        & 2.01    & 1.57     &-       & 1.57     &6.21    &2.42    & 1.72  \\
                    & Params (M)   &-        & 4.24    & 0.31     &-       & 0.06     &1.12    &0.18    & 0.13  \\
                    & FLOPs (M)    &-        & 45.71   & 41.78    &-       & 42.73    &60.70   &52.27   & 25.76  \\ \bottomrule[0.8pt] \hline
\end{tabular}
}
\end{table}
\section{CONCLUSION}
\label{CONCLUSION}
In this letter, we introduce a dual-branch residual network framework for cross-domain few-shot HSI classification.
This architecture is designed to fully exploit HSI spatial-spectral information by independently extracting, which are fused into comprehensive joint representations. To obtain the refined prototypes in few-shot scenarios, a contrastive learning constraint to reduce both the intra-class variation and inter-class similarity, improves the quality of the prototypes. Cross-domain adaptation is achieved by employing kernel space probability distribution aligning. Experimental results demonstrate the robust generalization and efficient classification of the proposed method on four public datasets.

\bibliographystyle{IEEEtran}
\bibliography{reference.bib}

\vfill
\end{document}